# Consensus-Threshold Criterion for Offline Signature Verification using Convolutional Neural Network Learned Representations

Paul Brimoh[1], Chollette C. Olisah[2*], Member IEEE

[1]Baze University, Abuja Nigeria.
[2]School of Engineering, University of the West of England, Frenchay, England, UK.

Abstract

*A genuine signer's signature is naturally unstable even at short time-intervals whereas, expert forgers always try to perfectly mimic a genuine signer's signature. This presents a challenge which puts a genuine signer at risk of being denied access, while a forge signer is granted access. The implication is a high false acceptance rate (FAR) – the percentage of forge signature classified as belonging to a genuine class. Existing work have only scratched the surface of signature verification because the misclassification error remains high. In this paper, a consensus-threshold distance-based classifier criterion is proposed for offline writer-dependent signature verification. Using features extracted from SigNet and SigNet-F deep convolutional neural network models, the proposed classifier minimizes FAR. This is demonstrated via experiments on four datasets: GPDS-300, MCYT, CEDAR and Brazilian PUC-PR datasets. On GPDS-300, the consensus threshold classifier improves the state-of-the-art performance by achieving a 1.27% FAR compared to 8.73% and 17.31% recorded in literature. This performance is consistent across other datasets and guarantees that the risk of imposters gaining access to sensitive documents or transactions is minimal.*

Keywords: Offline signature verification, CNN features, consensus-threshold criterion,

## 1. Introduction

The use of signature as a mode of authentication dates to ancient times and it is still widely adopted by numerous organizations, especially the financial institutions. Statistics from the United States (US) National Check Fraud Center[1] reveal that a loss of over $10 billion is recorded from forged cheques in the US alone on yearly basis. Therefore, the importance of signature verification systems cannot be overemphasized.

The human signature is characterized as a behavioral biometric because the neuromuscular system [1] is involved in signature inscription. The neuromuscular system is made-up of very large amounts of neurons and muscle fibers which control the pattern the hand makes as humans scribble their signature. In essence, the physical and emotional states of a signer can affect the stability of a signer's signature, even at short time intervals [2].

In signature verification, a good feature extraction algorithm is important [3] but it should be capable of capturing even the most miniature of pattern changes common of signatures of the same person referred to as the genuine signature. The variation that exists for this category of signature is termed the intra-signature variation. On the contrary, is the forged signature which is generated when a forger tries to impersonate a genuine signer's signature. This category of signature introduces the variation that exist of genuine and forged signature pairs termed inter-signature variation. For several decades, signature verification systems depended on handcrafted features. Discrete Wavelet Transform [4], Discrete Cosine Transform [5], Histogram of Oriented Gradients (HOG) [6] Local Binary Pattern [6], Gabor features [4], Scale Invariant Feature Transform (SIFT) [6], Gray-Level Co-occurrence Matrix (GLCM) [7] and Geometric features [8] and ensemble of local descriptors [9]. However, this category of features falls short in that they are unable to learn the miniature changes of a genuine signer's signature as it differs from its forged counterpart. Current approach is the adoption of learnable models for extracting signature patterns. However, there very few learnable models, like [15], and [17], and they have so far shown resilience against intra-person and inter-person variations than has been achieved with the handcrafted features [10], [11], [12], [13], and [14]. In this work, we pay particular attention to [15]. This is for the fact that two learnable models are considered: deep convolutional Neural Network (DCNN) signature models learned using genuine signatures only, and genuine signatures and forgeries; both termed SigNet and SigNet-F, respectively. The latter enables the model to update its knowledge of a genuine set given forgery. The good part is that this writer-independent feature approach is very applicable in the real world. By writer-independent feature it is meant that a model is trained to separate each signer in the learned representation space [16]. However, the writer-dependent classifier approach achieved a state-of-the-art result but included random forgeries as negative in the classifier training phase. The random forgeries are genuine signatures from other users. Meanwhile, it is almost impossible to have two persons with the same signature pattern. Therefore, a writer-dependent approach that determines the separation margin of a forgery and genuine sample from random forgeries is unrealistic and might be far from application in the real-world environment.

---
[1] http://www.ckfraud.org/statistics.html

Overall, the writer-dependent classifiers commonly employed in literature are: support vector machines (SVM) with linear kernel [15], [17], and [3]; with RBF kernel are [15], and [18]; then with hidden Marcov models (HMMs) and distance classifiers (Euclidean, Cosine, City-block and Mahalanobis) [19]; Naïve Bayesian; or the Ensemble of Classifiers [16], and [20]. For a detailed discussion on their success stories, readers are directed to [21]. Despite their successes, there is usually the problem of high misclassification error. We focus on SVM and distance classifiers because they are most used for learned models. The SVM(s) is known to have an outstanding performance in other areas of classification task such as face, palmprint, iris, and fingerprint as depicted in [22]. For signature classification, it struggles which may because the expert forgers always try to perfectly mimic a genuine signer's signature even more than the genuine signer. therefore, if addressed from a *one-class* point of view as shown in [18], there is a high likelihood that the SVM will assign good number of the forged probe signature samples to the genuine class. The same applies to the *two-class* SVM classifier where the genuine signature samples of other signers' act as forge class for a genuine signature sample of a signer. The initial trial-and-error experiment we carried out without kernel tuning showed that misclassification error can be reduced drastically if a forge set, no matter how small, is introduced during the classifier training. Otherwise, the misclassification error remains high, like in [15]. It is important to note that for the one-class or two-class SVM to achieve good performance accuracy, the kernel needs to be empirically determined and fine-tuned [23]. This is highly dependent on sample data distribution and known prior.

On the other hand, the distance classifiers, Euclidean, or cosine dissimilarity classification algorithms have their fair share of misclassification error. This is due to the sensitivity of distance classifiers to irrelevant features. However, we hypothesize that the misclassification error of these type of classifiers in the task of signature classification mainly hinges on the thresholding criterion. This criterion is critical in making a decision to assign a claimed identity to a genuine or forge class. Invariably, if the thresholding criterion is defined around a margin that somewhat separates the forge and genuine class, misclassification error can be minimized. Literature reveal that such line of thought has been treaded in very recent time by Alaei et al. [23]. In their work, they defined an adaptive dissimilarity margin termed the acceptance threshold which was computed from the confidence value per signer. This is an interesting writer-dependent classifier approach for separating a forge set from a genuine set. However, the genuine and forge set were used to tune the confidence value parameter for acceptance/rejection threshold determination. In the same vein Manjunatha et al. [24] redefined the margin in [24] by considering the confidence bound from the mean and standard deviation of the signature class.

We consider the following research questions to be relevant to the design of a real-world signature verification system.
- Which of the feature extraction methods is better-off at detecting the most minute of changes between a genuine and forged signature sets while extracting the behavioral patterns of a signature?
- Does a margin exist between genuine and forge signatures?
- Can this margin be explored for discriminating between forge and genuine signatures?

To address these questions, this paper proposes a framework that combines writer-independent features and writer-dependent classifier. We explore deep convolutional neural network models, SigNet and Signet-F [15] that learns writer-independent feature space and extracts discriminative signature cues unique to a single signer. We propose a writer-dependent classifier approach that takes into consideration only the genuine signatures for determining what we term a consensus-threshold which significantly minimizes false acceptance rate of forgeries while still giving access to a genuine signer. The remaining sections of this paper are organized as follows. The nature of signature is discussed in Section 2. In Section 3, the proposed methodology is detailed. The results and discussions are provided in Section 4. Lastly, conclusions are presented in Section 5.

2. The Signature

Like other biometrics, the signature serves as a medium for authenticating a claimed identity. To discuss the signature in-depth, lets walk through a typical application scenario of its use: The bank, for reasons of financial risk, collects several samples of signature of a customer on engagement for any of its services. The signatures become a true representation of the identity of the customer used over time for authentication. Now, suppose the customer makes a request, at some later dates, such as withdrawal above daily allowable limit or cashing cheques, his/her identity must have to be verified. If peradventure the newly provided signature sample of the customer differs greatly, access to requested services is denied. The forgery expert keeps this in mind and most often can come up with a signature that fits the true representation of a given customer's signature. Invariably, he/she has a better chance of been granted access than a genuine signer.

It has been well established that a genuine signature exhibit variability over time, a highlight can be seen in [25]. However, the extent of the close similarity shared between genuine signatures with forge signatures need to be stressed and quantified. The reason is this: since the expert forger tries as much to replicate a true signer's signature, there is the likelihood of having low variance between samples of a forge and genuine signature of a single signer. We demonstrate in Fig. 1(a) that despite feature extraction with a learned model, a close similarity is still shared between genuine and forge signatures, especially with SigNet-F. Further, in Table 1, a low variance can clearly be observed between the forge samples and the genuine samples. Also, a significant observation is that the mean and standard deviation of the forge samples, for almost all the dataset, is higher than the genuine samples. Though the difference is slight, it might be a significant knowledge for a classifier.

Evidently, the focus of a signature verification system should not lie on the feature extractor alone, the classifier is as well important. We consider an approach where a set of genuine signatures can be used to define a margin which can as much as it may be possible be used to separate a forgery signature from the genuine set. This is discussed further in the succeeding section.

## 3. Methodology

### 3.1. The Proposed Classifier Approach

Let a signature be described by n-dimensional features expressed as $\vec{x}$.

$$\vec{x} = [x^1, \cdots, x^n] \in X \quad (1)$$

where $X \subseteq R^n$ is a subset of the n-dimensional learned feature space.

For the learned feature space, we prefer the CNN based features of which SigNet and SigNet-F are two models that have achieved state of the art verification in literature. We split the genuine signature samples of a single signer into gallery set, $\{\vec{x}_g\}^G$, and probe set, $\{\vec{x}_p\}^G$, where $\vec{x}_g, \vec{x}_p \in X$, $g, p = \{1, \cdots, k\}$ and $k$ is the $k^{th}$ number of samples of a single signer belonging to the gallery or probe set. Now, we are interested in creating what we term the consensus signatures. Each single signer's gallery set is split into two unequal ratios; $\vec{x}_{g_a}$ and $\vec{x}_{g_b}$ with $a$ having the highest ratio than $b$. Further, the metric of $\vec{x}_{g_b}$ is computed against $\vec{x}_{g_a}$ and each metric value is compared against the centroid of their computed distances to yield the consensus signatures. The process of determining the consensus signature is described in Algorithm 1.

Having determined the consensus signature set, it is expected that these signatures are to some extent a true example, a sort of prototypes, of a given signer's identity.

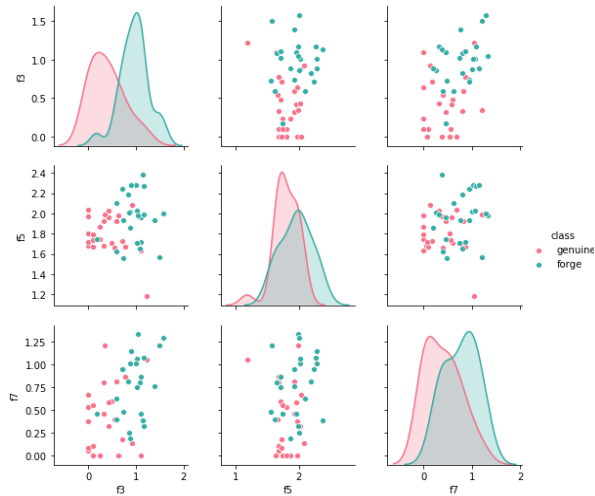

Figure 1: Comparing the statistical distribution of genuine (24) and forge (25) signatures using selected features on GPDS dataset. These features (f3, f5, and f7) were chosen for visualization as they are the only ones to show separation between the classes. The visualization is via *pairplot* with Seaborn.

Table 1 Comparing the mean and standard deviation of a single signer's genuine and forge signatures. Only the first signature samples, genuine and forge, of each dataset is used.

| Data | Feature | Measure | Genuine | Forge | Difference |
|---|---|---|---|---|---|
| CEDAR | SigNet | Mean | 0.6089 | 0.7583 | 0.1494 |
| | SigNetf | | 0.6381 | 0.7204 | 0.0823 |
| | SigNet | Standard Deviation | 0.1402 | 0.1540 | 0.0138 |
| | SigNetf | | 0.1817 | 0.1617 | 0.0200 |
| GPDS Signature 960 Grayscale | SigNet | Mean | 0.2751 | 0.2914 | 0.0163 |
| | SigNetf | | 0.2857 | 0.3622 | 0.0765 |
| | SigNet | Standard Deviation | 0.1532 | 0.1758 | 0.0226 |
| | SigNetf | | 0.1751 | 0.2143 | 0.0392 |
| BRAZILIAN (PUC-PR) | SigNet | Mean | 0.3357 | 0.4820 | 0.1463 |
| | SigNetf | | 0.5826 | 0.6044 | 0.0218 |
| | SigNet | Standard Deviation | 0.1378 | 0.2449 | 0.1071 |
| | SigNetf | | 0.1668 | 0.2856 | 0.1188 |
| MCYT-75 | SigNet | mean | 0.2887 | 0.2929 | 0.0042 |
| | SigNetf | | 0.2857 | 0.3622 | 0.0765 |
| | SigNet | Standard Deviation | 0.1941 | 0.1915 | 0.0026 |
| | SigNetf | | 0.1644 | 0.2030 | 0.0386 |

Therefore, of importance is the interest to minimize access gain for forge signatures as much as possible while still maintaining a reasonable access gain for genuine signatures. To achieve this, a threshold criterion is computed from the consensus signature of a single signer. The process is as described in Algorithm 2.

The classifier we have described thus far is writer-dependent. Let the consensus-threshold classifier function on a feature be denoted as $F(\vec{x})$. A new instance vector, $\vec{x}_{p_1}$, of a probe set $\{\vec{x}_p\}^G / \{\vec{x}_p\}^F$, to be classified as belonging to a genuine class or a forge class by a consensus-threshold classifier function $F(\vec{x})$, is determined based on whether the output of $F(\vec{x}_{p_1})$ is 1 (genuine) or 0 (forge), respectively. The classification process is as expressed:

$$F_{\tau_c}(\vec{x}_p) = \begin{cases} 1, & \text{if } F(\vec{x}_p) \geq \tau_c \\ 0, & \text{otherwise} \end{cases} \quad (2)$$

where $\tau_c$ is the consensus-threshold criterion.

---

**Algorithm 1** Determining consensus signature set

**Input**: metric matrix input $d_i = d_{\cos}(\vec{x}_{g_a}, \vec{x}_{g_b})$, $i = \{1, \cdots, I\}$

**Parameter**: $e$, a scalar to tune the centroid.

**Output**: metric matrix of consensus signatures, $\{\breve{d}_i, i = 1, \cdots, I\}$

Step 1: calculate mean of metric matrix, $\mu = \dfrac{1}{m}\sum_{i=1}^{m} d_i$

Step 2: $\mu' = \mu - e$, where $e = e^{-4}$

Step 3: retrieve matrix $d$ of size r x c;
 for $u \leftarrow$ 1 to $r$ do
  for $v \leftarrow$ 1 to $c$ do
   if $d(u,v) \geq \mu'$ then
    $\breve{d}_i \leftarrow d(u,v)$
   endif
  endfor
 endfor

---

As earlier stated, the consensus threshold function, $F(\vec{x})$, seeks to minimize FAR as much as possible. By generating the consensus signature set of a single signer (as described in Algorithm 1), it will be loosely possible for a forge set to be classified as genuine. This can be demonstrated using the set theory difference operation. Given genuine and forge sets, an instance belonging to a genuine set can be distinguished from a forge set and vice versa as diagrammatically illustrated in Fig. 2 and expressed as follows:

---

**Algorithm 2** Compute Consensus-Threshold

**Input**: metric matrix of consensus signatures, $\{\breve{d}_i, i = 1, \cdots, I\}$

**Parameter**: $\alpha$, significance value; $e$, a scaler to tune the bound of the threshold where $e = e^{-4.5}$.

**Output**: consensus threshold, $\tau_c$

Step 1: calculate mean of consensus signatures matrix,

$$\mu = \frac{1}{m}\sum_{i=1}^{m}\breve{d}_i$$

Step 2: calculate standard deviation (*std*) of consensus signature matrix,

$$\sigma = std(\breve{d}_i)$$

Step 3: with number of elements in the matrix, m, calculate confidence value, $f$

$$f = (\sigma/\sqrt{m}) \times \alpha$$

Step 4: compute the consensus threshold, $\tau_c$

$$\tau_c = (\mu - f) - e$$

---

Let $\{\vec{x}_{g,p}\}^G = G; \{\vec{x}_p\}^F = F$   (3)

G-F $= \{x | x \in G \text{ and } x \in \bar{F}\} = G \cap \bar{F}$   (4)

Similarly,

F-G $= \{x | x \in F \text{ and } x \in \bar{G}\} = F \cap \bar{G}$   (5)

By (4) it is clear that some genuine instances might be misclassified as forge samples. Though the risk is minimal, and for cases as this, other measures can be used. For example: a transaction requiring signature can be authorized given that three or more signature samples of a signer can be verified.

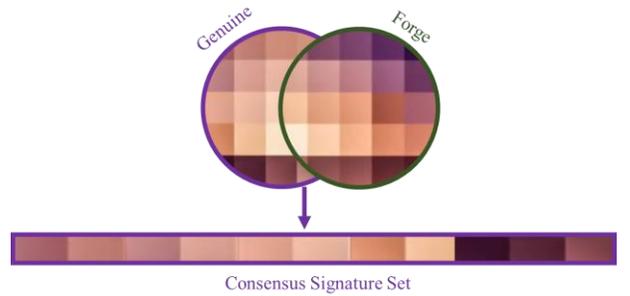

Figure 2: Demonstrating consensus signature generation using sets

### 3.2. The Experimental Protocol

This paper utilizes the popularly used signature datasets GPDS-960 [26], MCYT-75 [27], CEDAR [28] and the Brazilian PUC-PR [29] with features extracted using the deep CNN features, termed SigNet and SigNet-F designed by Hafemann et al. [15]. There are 2048 feature vectors obtained from the last layer (FC7 layer) of the deep CNN for representing a signer's signature. According to [15] the SigNet learns using genuine signatures only and SigNet-F learns using genuine signatures and forgeries. The feature sets are as available[2], except for the GPDS dataset features, only the exploitation set, about 300 signers, is used. The logistics of signers as per dataset is summarized in Table 2. We setup the experiment scenario a lot differently from how it is conventionally done. We did not create random samples. This is for the fact that the argument of a person impersonating another individual by signing his/her signature is not logical in the financial sector, and undermines the treat factor introduced by forgery.

Considering that the output of $F(\vec{x})$ can be of genuine (1) or forge (0), the following possibilities are illustrated in what is statistically termed the confusion matrix: (i) True Positives (TP): number of genuine signatures correctly classified as belonging to the genuine class; (ii) True Negatives (TN): number of forge signatures correctly classified as belonging to the forge class; (iii) False Positives (FP): number of forge signatures misclassified as belonging to the genuine class; (iv) False Negatives (FN): number of genuine signatures misclassified as belonging to the forge class. On the basis of the confusion matrix the correctness and accuracy of the model can be evaluated using statistical measures of performance such as classification accuracy, false acceptance rate (FAR), false rejection rate (FRR), and average error rate (AER) [30].

The classification accuracy is the number of overall correct predictions $F(\vec{x})$ makes, FRR is the portion of times that $F(\vec{x})$ fails to predict a genuine signature as belonging to a genuine class, FAR is the portion of times $F(\vec{x})$ predicts that a forge signature is of a genuine class, and AER is the average of FAR and FRR.

Table 2 Summary of signature datasets

| Dataset | Signers | Genuine | Forgery | Total |
| --- | --- | --- | --- | --- |
| CEDAR | 55 | 24 | 20 | 2420 |
| GPDS Signature 960 Grayscale | 300 | 24 | 25 | 14700 |
| BRAZILIAN (PUC-PR) | 60 | 40 | 20 | 3600 |
| MCYT-75 | 75 | 15 | 15 | 2250 |

4. Results and Discussion

The signature verification system performance of the proposed consensus threshold classifier for a deep CNN learned offline signature features is reported. The empirical analysis of threshold parameters using the proposed Consensus-Threshold Criterion which is a distance-based classifier, and it uses the cosine function for computing distance. The threshold value of 99.999% (the significance value) is shown to achieve the best performance. Its justification comes from the analysis in [33] which reveals that in cases of high risk like in financial institutions, such significance value should be preferred. Using the best performing parameter for experiments, results are discussed from three perspectives: (i) Analysis of performance using different signature datasets; (ii) consensus threshold classifier performance against Min, Max, Mean, and confidence interval-valued thresholding; and (iii) comparison with the state-of-the-art.

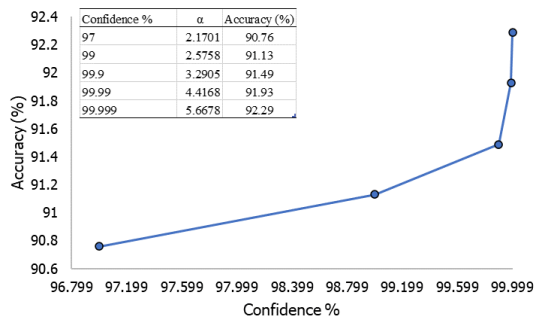

Figure 3: Performance effect of varying parameter $\alpha$ on CEDAR dataset.

4.1. Analysis of performance using different signature datasets

In this experiment, SigNet and SigNet-F are used as feature extraction models and classifier decision is based on the proposed consensus-threshold criterion. Overall, it is observed from Table 3 that the performance of the proposed verification system is in accordance with the goal of the paper. The FAR is substantially minimized as we hypothesized and it is a desired performance for signature systems. A FAR of 1.16, 1.27%, 5.00%, 10.2% is achieved for BRAZILLIAN-75, GPDS-300, CEDAR, and MCYT, respectively. Just like in [23] where statistical measures are used within the classifier, the number of samples is observed to influence performance. This can be seen for the FAR obtained on MCYT dataset with very few reference genuine signer's signatures. In comparison, the FAR obtained on MCYT dataset is above 7% increase to the most FAR obtained for GPDS-300, CEDAR, or BRAZILLIAN-75 datasets where more reference samples are available. Further, the use of genuine and forge signatures to learn the discriminative space of signatures by Hafemann et al. 2017 [15] show to be significant to our signature verification system. Also, it can be observed that SigNet-F improved the FAR of the signature verification system by 2.41%, 1.24% for GPDS-300, and Brazillian-75, respectively, over the SigNet model.

---

[2] https://github.com/luizgh/sigver_wiwd

Table 3 Performance of consensus threshold classifier on GPDS-300, CEDAR, BRAZILIAN (PUC-PR) and MCYT datasets

| Dataset | Feature | Gallery | | Probe | | Accuracy (%) | FAR (%) | FRR (%) | AER (%) |
|---|---|---|---|---|---|---|---|---|---|
| | | Genuine | | | Forge | | | | |
| | | $\{\vec{x}_{g_a}\}^G$ | $\{\vec{x}_{g_b}\}^G$ | $\{\vec{x}_p\}^G$ | $\{\vec{x}_p\}^F$ | | | | |
| GPDS-300 | SigNet | 14 | 5 | 5 | 25 | 93.91 | 3.68 | 18.27 | 10.97 |
| | SigNet-f | | | | | **96.06** | **1.27** | 17.33 | 9.30 |
| CEDAR | SigNet | 14 | 5 | 5 | 20 | 91.49 | **5.00** | 22.54 | 13.77 |
| | SigNet-f | | | | | **92.29** | 5.27 | 17.45 | 11.36 |
| BRAZILIAN (PUC-PR) | SigNet | 25 | 10 | 5 | 20 | 94.40 | 2.40 | 18.00 | 10.20 |
| | SigNet-f | | | | | **95.40** | **1.16** | 18.67 | 9.91 |
| MCYT | SigNet | 5 | 5 | 5 | 15 | **85.95** | **10.27** | 25.41 | 17.84 |
| | SigNet-f | | | | | 85.41 | 12.70 | 22.37 | 17.53 |

Table 4 Evaluation of different threshold functions on EDAR dataset (rates in %).

| Function | Feature | Accuracy | FAR | FRR | AER |
|---|---|---|---|---|---|
| Max-value | SigNet | 80.00 | 0.00 | 100 | 50.00 |
| | SigNet-F | 96.06 | 0.00 | 100 | 50.00 |
| Min-Value | SigNet | 79.27 | 20.45 | 21.82 | 21.35 |
| | SigNet-F | 81.45 | 18.18 | 20.20 | 19.19 |
| Mean-value | SigNet | 88.29 | 13.82 | 55.27 | 34.54 |
| | SigNet-F | 88.29 | 13.82 | 55.27 | 34.54 |
| **Confidence Interval Threshold [24]** | SigNet | 90.40 | 9.27 | 10.91 | 10.09 |
| | **SigNet-F** | 90.40 | 9.27 | 10.91 | 10.09 |
| Confidence Interval Threshold-a [23] | SigNet | 46.84 | 41.45 | 100 | 70.73 |
| | SigNet-F | 46.62 | 41.82 | 99.64 | 70.73 |
| **Confidence Interval Threshold-b [23]** | SigNet | 52.87 | 58.91 | 0 | 29.46 |
| | **SigNet-F** | 54.84 | 56.27 | 0.36 | 28.32 |
| **Proposed** | **SigNet** | 91.49 | **5.00** | 22.54 | 13.77 |
| | **SigNet-F** | 92.29 | 5.27 | 17.45 | 11.36 |

Confidence Interval Thresholding-a: 0 is genuine, 1 is forge; Confidence Interval Thresholding-b: 0 is forge, 1 genuine

### 4.2. Consensus threshold distance classifier against other measures

This experiment is carried out using the CEDAR dataset with features extracted using the SigNet and SigNet-F models. In other to determine the performance validity of the proposed consensus threshold distance classifier, other distance classifier approaches are evaluated as well. These approaches are the max-value, min-value, mean-value, confidence interval-value (shortened as interval-value).

From the experimental results in Table 4, it can be observed that the best performing distance classifier approach, is the consensus threshold. It achieved a FAR of 5.00%. Though interval-valued approach [24] achieved a lower AER, its FAR of 9.27% is higher by 4.27% compared to the proposed. For a signature verification system, its performance is highly dependent on FAR; the lower the FAR, the less likely for imposters to gain access to risk sensitive transactions like debiting an account via cheques in financial institutions.

### 4.3. Comparison with the state-of-the-art

Results of the state-of-the-art on the Brazillian (PUC-PR) and GPDS-300 datasets are presented on Table 5. We report just-as-it-is in literature the results of different researchers despite varying sizes used for these datasets.

On Brazillian dataset, as presented in literature, the proposed classifier with SigNet feature model achieves 2.40%, a performance difference of 3.55% and 4.77 in terms of FAR when compared against [31] and [15] who used the same SigNet feature model. However, classification in [31] is from a writer-independent approach while [15] is from a writer-dependent approach. In comparison with other methods (feature extraction and classifier), our proposed method is observed to have low tolerance for forgery. Further, the proposed method reached as low as 1.16% FAR with SigNet-F feature model.

Table 5 Comparison with the state-of-the-art on BRAZILIAN (PUC-PR) and GPDS-300 datasets (rates in %).

| Reference | Features | Classifier | #Samples | FAR | FRR | AER |
|---|---|---|---|---|---|---|
| BRAZILIAN (PUC-PR) | | | | | | |
| Bertolini *et al.* [16] | Graphometric | Ensemble of Classifiers | 15 | 6.48 | 10.16 | 8.32 |
| Batista *et al.* [3] | Pixel density | HMM & SVM | 30 | 13.5 | 7.5 | 10.5 |
| Rivard *et al.* [20] | Extended-Shadow Code (ESC) & Directional Probability Density Function (DPDF) | Ensemble of Classifiers | 15 | 11.15 | 11 | 22.15 |
| Eskandar *et al.* [32] | Extended-Shadow Code & Directional Probability Density Function | WD Thresholding | 30 | 13.5 | 7.83 | 10.66 |
| Hafemann, Sabourin & Oliveira [17] | Deep CNN | WD SVM (Linear) | - | 27.17 | 1.00 | 14.09 |
| **Hafemann *et al.* [15]** | **SigNet** | **WD SVM (Linear)** | 5 | 7.17 | 4.63 | 5.9 |
| | | | 15 | 10.70 | 1.22 | 11.92 |
| | | | 30 | 12.62 | 0.23 | 6.43 |
| **Souza, Oliveira & Sabourin [31]** | **SigNet** | **WI SVM (Linear)** | 5 | 5.95 | 5.95 | 5.95 |
| | | | 15 | 5.13 | 5.13 | 5.13 |
| | | | 30 | 4.90 | 4.90 | 4.90 |
| Proposed | SigNet | WD Consensus Thresholding | 10 | 2.40 | 18.00 | 10.20 |
| **Proposed** | **SigNet-F** | **WD Consensus Thresholding** | **10** | **1.16** | 18.67 | 9.91 |
| GPDS-300 | | | | | | |
| Bouamra *et al.* [18] | Run-Length Based Feature | WD SVM (Linear) | 4 | 29.50 | 28.10 | 28.80 |
| | | | 8 | 28.59 | 28.61 | 28.60 |
| | | | 12 | 28.32 | 30.20 | 58.52 |
| Hafemann, Sabourin & Oliveira [17] | Deep CNN | SVM (Linear) | 12 | 25.58 | 11.93 | 18.76 |
| Ribeiro *et al.* [33] | Deep Belief Networks | WD SVM | - | 14.67 | 20.25 | 17.46 |
| Alaei *et al.* [23] | Interval Symbolic Representation | WD Confidence Interval-Thresholding | 5 | 17.31 | 17.22 | 17.27 |
| | | | 12 | 7.90 | 14.76 | 11.33 |
| Eskandar *et al.* [32] | (ESC) & (DPDF) | WD Thresholding | 10 | 10.64 | 17.43 | 14.04 |
| **Hafemann *et al.* [15]** | **SigNet-F** | **WD SVM (Linear)** | **5** | **8.18** | **9.28** | **8.73** |
| | | | **12** | **6.16** | **6.80** | **6.48** |
| Proposed | SigNet | WD Consensus Thresholding | 5 | 3.68 | 18.27 | 10.97 |
| **Proposed** | **SigNet-F** | **WD Consensus Thresholding** | **5** | **1.27** | 17.33 | 9.30 |

On GPDS-300, the proposed method with SigNet-F achieved FAR of 1.27%, which from observation is the best performing method for the GPDS-300 dataset. This is 6.91% less when compared against [15]. Though the AER of [15] is low, 8.73% with 5 genuine reference signatures in the training set, it is just less by 0.57% in comparison to our method. Also, our proposed method by far outperformed the method in [23], a 16.04% difference. Even though this method shares similar statistical concept as ours, it still falls short with respect to performance.

A general observation is that the FAR and FRR of the state-of-the-art results perform almost equally for all the methods in literature. This implies that the state-of the-art methods allow, at high percentages, forge signatures to claim identity of a genuine signature. However, the opposite is the case in real-world where forgeries result in huge risk to financial institution. Therefore, it has become a performance burden that signature systems will have to overcome. Though our proposed method overcomes this burden at the expense of genuine signatures, it is still a plus because a genuine signer can provide more signature samples at the point of verification, which is likely to increase a genuine signature's chances of acceptance. This can be made a status-quo in signature verification systems because it will help to curb forgery. However, FRR and AER can alongside FAR be minimized if the classifier is exploited when training the SigNet(s).

5. Conclusion

In this paper, we presented a consensus-threshold distance-based classifier criterion for offline writer-dependent signature verification. Using features extracted from SigNet and SigNet-F deep convolutional neural network models, the proposed classifier was able to minimize FAR on the four datasets: GPDS-300, MCYT, CEDAR and Brazilian PUC-PR datasets. Particularly, on GPDS-300, the consensus threshold classifier improved the state-of-the-art performance by achieving 1.27% FAR compared to 8.73% and 17.31% in literature. This is a guarantee that the risk of forgery signer gaining access to sensitive documents can be minimized. However, with high FRR and AER, which burdens on the authentication of a genuine signer, developing an algorithmic framework that uses the consensus-threshold criterion for online signature verification is most likely to drastically minimize the FRR and AER errors. This consideration will be made in the future work.